%% file: acl2023.tex
\pdfoutput=1

\documentclass[11pt]{article}

\usepackage{ACL2023}
\usepackage{times}
\usepackage{latexsym}
\usepackage{fontawesome5}
\usepackage[T1]{fontenc}
\usepackage{inconsolata}
\usepackage{refstyle}
\usepackage{multirow}
\usepackage{times}
\usepackage{latexsym}
\usepackage[utf8]{inputenc}
\usepackage{xcolor,colortbl}
\usepackage{microtype}
\usepackage{url}
\usepackage{longtable}
\usepackage{tikz}
\usepackage{tabularx}
\usepackage{lscape}
\usepackage{hhline}
\usepackage{bbm}
\usepackage{array}
\usepackage{xspace}
\usepackage{graphicx}
\usepackage{booktabs}
\usepackage{makecell}
\usepackage{amsfonts}       
\usepackage{nicefrac}       
\usepackage{microtype}      
\usepackage{multirow}
\usepackage{caption}
\usepackage{amsmath}
\usepackage{float} 
\usepackage{ragged2e}
\usepackage{breqn}
\usepackage{soul}
\usepackage[utf8]{inputenc}

\usepackage{cleveref}
\crefformat{section}{\S#2#1#3}
\crefformat{subsection}{\S#2#1#3}
\crefformat{subsubsection}{\S#2#1#3}
\crefrangeformat{section}{\S#3#1#4 to~\S#5#2#6}
\crefmultiformat{section}{\S#2#1#3}{ and~\S#2#1#3}{, #2#1#3}{ and~#2#1#3}
\Crefformat{figure}{#2Fig.~#1#3}
\Crefmultiformat{figure}{Figs.~#2#1#3}{ and~#2#1#3}{, #2#1#3}{ and~#2#1#3}
\Crefformat{table}{#2Tab.~#1#3}
\Crefmultiformat{table}{Tabs.~#2#1#3}{ and~#2#1#3}{, #2#1#3}{ and~#2#1#3}
\Crefformat{appendix}{#2Appx.~\S#1#3}
\crefformat{algorithm}{Alg.~#2#1#3}
\Crefformat{equation}{#2Eq.~#1#3}

\newcommand{\stitle}[1]{\noindent{\bf #1.}}
\usepackage{microtype}

\usepackage{inconsolata}

\newcommand{\modelname}{\textsc{EFactSum}\xspace}

\newcommand{\usc}{\raisebox{5pt}{\includegraphics[height=12pt]{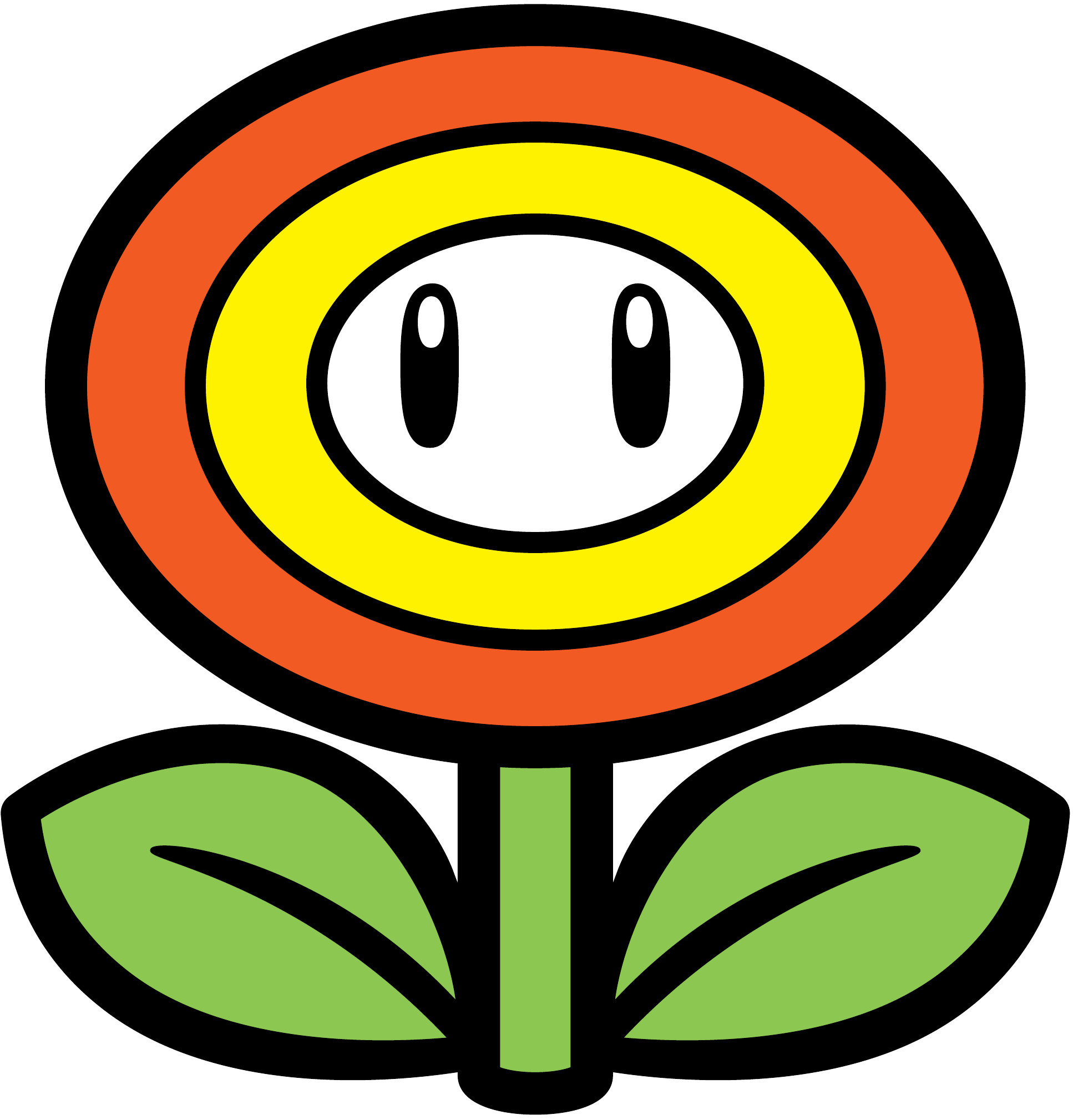}}}
\newcommand{\iitm}{\raisebox{5pt}{\includegraphics[height=12pt]{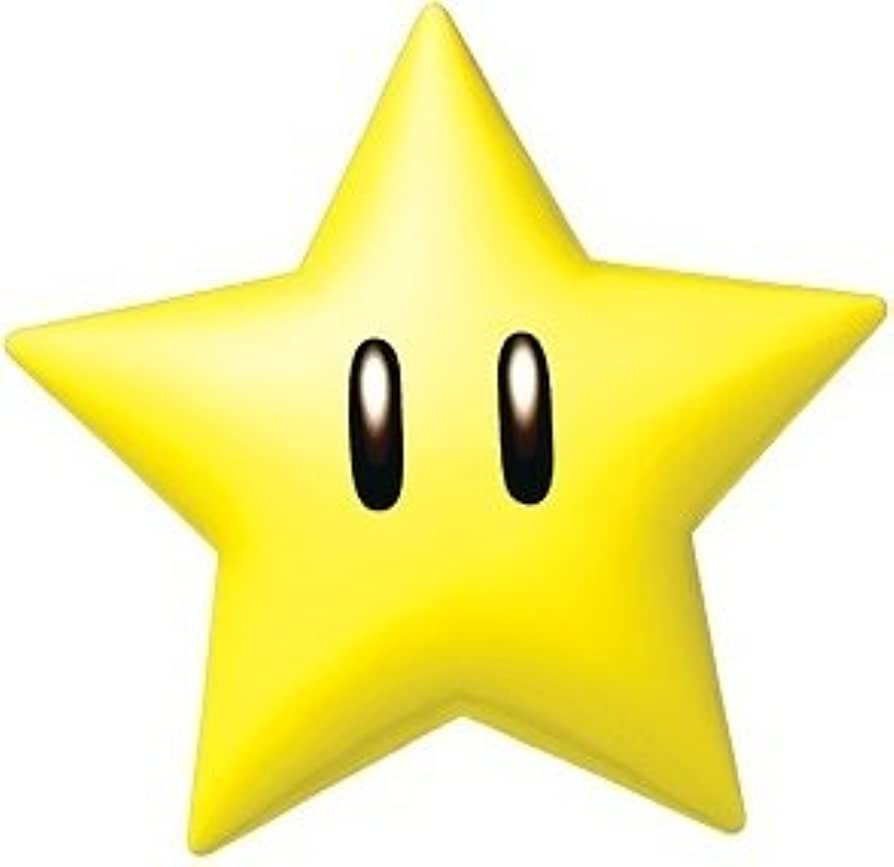}}}

\usepackage{times}
\usepackage{latexsym}

\usepackage[T1]{fontenc}

\usepackage[utf8]{inputenc}

\usepackage{microtype}

\usepackage{inconsolata}

\title{Improving Factuality of Abstractive Summarization \\
without Sacrificing Summary Quality }

\author{\textbf{Tanay Dixit}\iitm\thanks{\;\;This work was done when the first author was visiting the University of Southern California.} ~~~
        \textbf{Fei Wang}\usc  ~~~
        \textbf{Muhao Chen}\usc \\
        \iitm Indian Institute of Technology Madras ~~~
        \usc University of Southern California \\
        {\tt dixittanay@gmail.com} ~~
        {\tt \{fwang598,muhaoche\}@usc.edu}
        }

\begin{document}
\maketitle
\begin{abstract}

Improving factual consistency of abstractive summarization has been a widely studied topic. However, most of the prior works on training factuality-aware models have ignored the negative effect it has on summary quality. We propose \modelname (i.e., \textbf{E}ffective \textbf{Fact}ual \textbf{Sum}marization), a candidate summary generation and ranking technique to improve summary factuality without sacrificing summary quality. We show that using a contrastive learning framework with our refined candidate summaries leads to significant gains on both factuality and similarity-based metrics. Specifically, we propose a ranking strategy in which we effectively combine two metrics, thereby preventing any conflict during training. Models trained using our approach show up to 6 points of absolute improvement over the base model with respect to FactCC on XSUM and 11 points on CNN/DM, without negatively affecting either similarity-based metrics or absractiveness.\footnote{Code is available at \url{https://github.com/tanay2001/EFactSum}.}


\end{abstract}

\section{Introduction}
Although recent methods have made significant improvements in abstractive summarization \cite{lewis-etal-2020-bart, raffel2020exploring,10.5555/3524938.3525989}, they do still lack a very critical component - factual consistency. Recent works \citep{cao-etal-2020-factual, kryscinski-etal-2019-neural, maynez-etal-2020-faithfulness} have shown that a majority of the model-generated summaries are unfaithful and suffer from a wide range of hallucination \citep{tang2022understanding}. Making summarization models factually consistent is critical for its trustworthiness in real-world applications. 

Recent studies have made several attempts to improve factuality of abstractive summarization by either modifying the maximum likelihood estimation (MLE) training objective \citep{cao-wang-2021-cliff, goyal-durrett-2021-annotating}, directly optimizing factuality metrics using reinforcement learning \citep{cao-etal-2022-hallucinated} or improving the quality of the training data \citep{goyal-durrett-2021-annotating, nan-etal-2021-entity}. However, most of these works have reported a negative relationship between factual consistency and summary quality\footnote{summary quality as measured by metrics like ROUGE, BERTScore, etc.}. For example, \citet{goyal-durrett-2021-annotating} improve factuality at a cost of a 6-point drop in ROUGE-L, \citet{wan-bansal-2022-factpegasus} also observe a 2-point drop in ROUGE-L. Prior approaches have also optimized factuality at the cost of abstractiveness \cite{ladhak-etal-2022-faithful}. 
This leads to a critical question: \emph{Can we improve the factuality of summarization without the cost on the summary quality?}

\begin{figure}[t!]
    \centering
    \includegraphics[scale=0.37]{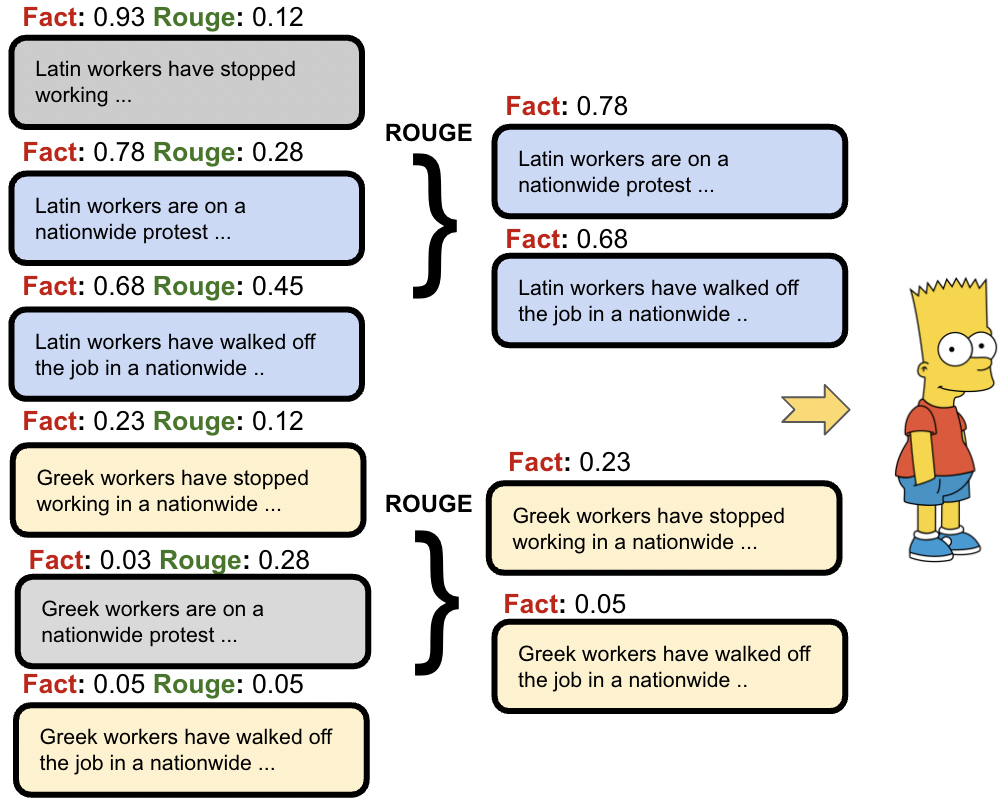}
    \caption{Overview of our approach. For a given article, we generate a number of summaries that can be either factual (blue) or non-factual (yellow). Grey summaries are filtered out. We select a balanced set using ROUGE and then finally train the model to rank them based on the factuality score.}
    \label{figure:teaser_fig}
\end{figure}

To this end, we propose \modelname (i.e. \textbf{E}ffective \textbf{Fact}ual \textbf{Sum}marization): A candidate summary generation and ranking technique for contrastive summarization training (\Cref{figure:teaser_fig}) that not only achieves significant gains in factuality of abstractive summarization but also improves the summary quality. 
Unlike prior works which often sacrifice summary quality for improving faithfulness,

we take an alternative approach to improve both faithfulness and summary quality. We make use of the fine-tuning strategy by \citet{liu-etal-2022-brio} and make key modifications to the ranking process. As depicted in \Cref{figure:teaser_fig} we start 
with generating a number of candidate summaries using existing fine-tuned models. Using these summaries, we select a subset by effectively combining two evaluation metrics of the two different criteria (\Cref{sec:ranking_strat}), thus avoiding optimizing one at the cost of the other. This technique helps obtain gains over methods that simply optimize one metric (\Cref{sec:ablation}). 
The promising results by \modelname on XSUM and CNN/DM have shown consistent improvements in both aspects over strong baselines, demonstrating effectively enhanced summarization factuality without sacrificing the quality.


\section{Approach}
\label{sec:approach}
Given a document ($D$), the task of summarization seeks to generate its summary ($S$) that satisfies some conditions like factuality, coherence, etc. The standard fine-tuning process involved the use of Maximum Likelihood Estimation (MLE). 
Inspired by \citet{liu-etal-2022-brio}, in addition to the cross-entropy loss, we incorporate a contrastive loss that encourages models to provide a higher probability mass to the more factual summaries. Formally, for every training document $D$ and a ranked list of the most probable candidate summaries $[S_{1}, S_{2},\dots S_{n}]$, the model learns to rank the summaries according to the factuality score. To achieve this, we make use of the following loss:

\begin{equation}
\label{eq:contr}
      \mathcal{L}_{CL} = \sum_i \sum_{j > i} \max(0, f(S_j) - f(S_i) + \lambda_{ij}) ,
\end{equation}

\noindent
where $S_i$ and $S_j$ are two different candidate summaries and $S_i$ ranks higher than $S_j$, 
$\lambda_{ij} = (j - i) * \lambda$ is a rank-based margin, 
and $f(.)$ is the estimated log-probability normalized by length:

\begin{equation}
    f(S) = \frac{\sum_{t=1}^{l} \log p_{g_\theta} (s_t |D, S_{<t}; \theta)}{|S|^\alpha} .
\label{eq:score}
\end{equation}

\stitle{Candidate Set Generation} To generate the candidate summarization set $\{S_{i}\}$, we make use of an existing model and sample summaries using beam search \cite{Vijayakumar_Cogswell_Selvaraju_Sun_Lee_Crandall_Batra_2018}. We observe that just using the model trained with cross-entropy leads to generating a number of unfaithful summaries. In order to generate more faithful summaries, we make use of factually improved models.

\stitle{Ranking Strategy} 
\label{sec:ranking_strat}
Since our primary goal is to optimize factuality without adversarially affecting summary quality, we need to consider two metrics while deciding the ideal ranking. In order to measure the factuality of $S_{i}$, we choose FactCC \citep{kryscinski-etal-2020-evaluating} 
because it correlates well with human judgments of faithfulness \cite{pagnoni-etal-2021-understanding} and 
it is also computationally more efficient than other question-answering based metrics \cite{scialom-etal-2021-questeval}. To measure the summary quality, we use the popular ROUGE metric \citep{lin-2004-rouge}. Now, amongst the set of candidate summaries 
that have been scored to be faithful, we further choose the top $m$ summaries that have the highest ROUGE score. 
We select the set of unfaithful summaries in the same way just that we choose the $m$ summaries with the lowest ROUGE scores. This technique of incorporating two evaluation metrics helps overcome the inherent conflict \cite{chaudhury-etal-2022-x}. We highlight the importance of the proposed steps in \cref{sec:ablation}.
At last, these $2m$ summaries are used in creating the ranked list of candidate summaries for each article in the training set. The intuition behind this approach is that since the FactCC scores are not confidence scores, summaries from only one set can not provide sufficient supervision signals. Instead, training the model with balanced summaries from both sets would be beneficial.

Finally, our training objective combines the cross-entropy loss and our contrastive loss
\begin{equation}
\label{eq:mul}
    \mathcal{L}_{total} = \mathcal{L}_{CE} + \gamma \mathcal{L}_{CL} ,
\end{equation}
where $\gamma$ is the weight of the contrastive loss.


\input{main_results}
\input{examples}

\section{Experiments}

We state the experimental setup in \Cref{sec:exp_setup} and report the results in \Cref{sec:main_results}, followed by an abstractiveness analysis in \Cref{sec:analysis}. In \Cref{sec:ablation}, we analyze the importance of the various components in our approach.
\subsection{Experimental Settings}
\label{sec:exp_setup}
\stitle{Datasets}
\label{sec:datasets}
To understand the effectiveness of \modelname, we make use of two widely-used news summarization datasets, XSUM \citep{narayan-etal-2018-dont} and CNN/DM \citep{HermannKGEKSB15}.

\stitle{Baselines}
\label{sec:baselines}
In addition to models fine-tuned using \textit{cross-entropy} and competitive fine-tuning techniques: \textbf{BRIO} \cite{liu-etal-2022-brio}, we compare \modelname with prior works that have modified the fine-tuning process to improve factuality, including (1) \textbf{CLIFF} \cite{cao-wang-2021-cliff} which uses contrastive learning to train summarization models to differentiate between consistent and hallucinated summaries, (2) \textbf{FASum} \cite{zhu-etal-2021-enhancing} that modifies the Transformer architecture by incorporating knowledge graphs for factual consistency, and (3) \textbf{DAE} \cite{goyal-durrett-2021-annotating} that masks out the nonfactual tokens during training. This comparison is only available for the XSUM dataset.

\stitle{Metrics}
\label{sec:metrics}
To evaluate factuality, we make use of FactCC \citep{kryscinski-etal-2020-evaluating}, a popular metric that uses a BERT-based metric to measure whether the generated output is faithful. We also consider DAE \citep{goyal-durrett-2020-evaluating}, a textual-entailment-based metric that correlates well with human judgment of factuality \citep{tang2022understanding}. 
It uses an arc entailment model to evaluate the factuality of a summary. We make use of the token-level score in order to complement the sentence-level scores from FactCC. For quality assessment, we use ROUGE \cite{lin-2004-rouge} and BERTScore \cite{zhang2019bertscore} to evaluate the summary against the reference.


\stitle{Implementation Details}
\label{sec:imp_details}
We use CLIFF and cross-entropy trained models to generate the candidate set of summaries ($S_{1}, S_{2}, ..., S_{n}$).
We  use $n =6$ and only retain those training articles that contain at least 2 factual and non-factual candidate summaries. Using this new subset of training data, we fine-tune BART-Large \cite{lewis-etal-2020-bart} on CNN/DM and PEGASUS \cite{10.5555/3524938.3525989} on XSUM. More details are in
\Cref{sec:app_imp_details}.


\subsection{Main Results}
\label{sec:main_results}
We report the results of the model fine-tuned using our approach in \Cref{tab:main_results}. Outputs of models fine-tuned using our strategy are presented in \Cref{tab:example} and \Cref{sec:app_outputs}. Overall we can observe the proposed \modelname leads to improvements on both the factuality metrics 
while preserving or improving the performance on reference-based similarity metrics.
\noindent\par For XSUM, \modelname achieves a notable relative gain of 25\% on FactCC and 3\% on DAE (token) in comparison to PEGASUS while simultaneously showing non-trivial gains on both ROUGE and BERTScore. Although \modelname is trained to optimize FactCC, it also does well on the other evaluation metric, thus pointing out that the training process does not exploit any biases related to the evaluation metrics. One should note that although CLIFF does better on DAE, it is sacrificing summary quality. A similar story holds for CNN/DM also where \modelname achieves a relative gain of 20\% and 16\% on FactCC and DAE respectively. Unlike some of the prior works, this gain in factuality has not come at a cost of summary quality or abstractivness (\Cref{sec:analysis}). Although BRIO outperforms our approach on ROUGE and BERTScore, it substantially decreases factuality score, which is not desirable. Our approach aims to strike a balance between factuality and summary quality.


\begin{figure}[t!]
    \centering
    \includegraphics[scale=.37]{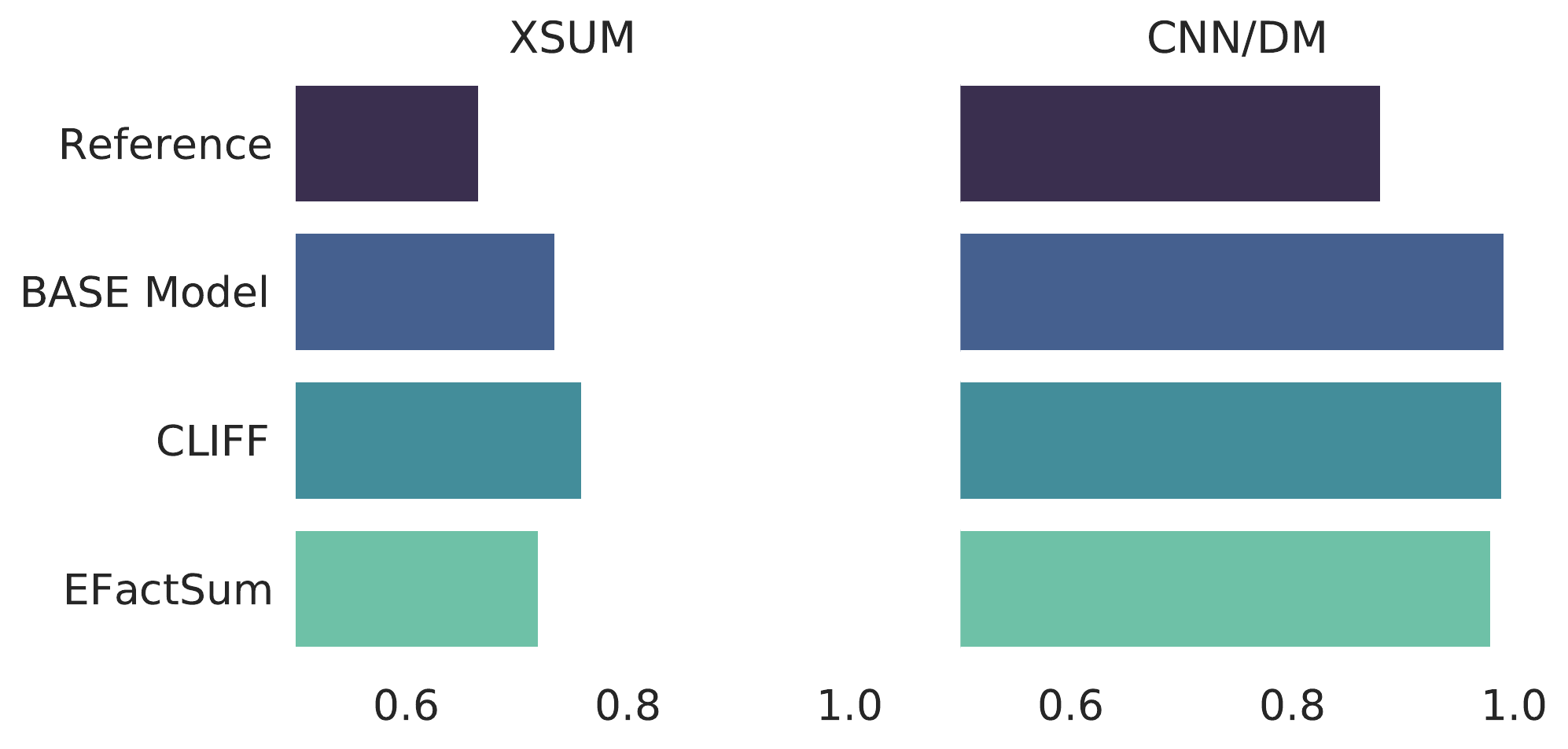}
    \caption{Extractiveness scores for the various models on both the datasets. The x-axis represents the Extractiveness calculated using the \textit{coverage} score defined by \citet{grusky-etal-2018-newsroom}. Smaller the extractivness score, more the abstraction in the summaries.}
    \label{fig:tradeoff}
\end{figure}


\subsection{Factuality vs Abstractiveness Tradeoff}
\label{sec:analysis}

\noindent\citet{ladhak-etal-2022-faithful} show that it is naively possible to increase the factuality of generated summaries by increasing extractiveness (decreasing abstractiveness). Hence we analyze the extractiveness level of the generated summaries to understand if our method suffers from this tradeoff. Along with the extractiveness scores \cite{grusky-etal-2018-newsroom}, we compute the MINT (Metric for lexical INdependence of generated Text) scores and the abstractiveness-adjusted metrics scores \cite{dreyer-etal-2023-evaluating}. \Cref{fig:tradeoff} depicts the extractiveness levels for the various summarization systems. Scores are also presented in \Cref{sec:app_extr_results}. We can observe that the extractiveness score for our model (\modelname) is lesser than other models; it also achieves higher MINT scores (\Cref{tab:mint_scores}), which measures the abstractiveness of the summaries. Additionally, \modelname shows higher scores for abstractiveness calibrated FactCC metric ($\mu$FactCC) for both datasets. This clarifies that the additional gains in factuality are not at a cost of absractiveness. 

\begin{table}[t]
    \centering
    \begin{tabular}{llcc}
    \toprule
    \bf Dataset  & \bf Model & \bf MINT  & \bf $\mu$FactCC \\
    \midrule
        \multirow{3}{*}{CNN/DM} & BART & 57.94 & 42.14 \\
        & CLIFF & 52.18 & 39.77 \\
         & \modelname & \bf 60.70 & \bf 47.47 \\
    \midrule
            \multirow{3}{*}{XSUM} & PEGASUS & 25.21 & 44.12 \\
        & CLIFF &  25.31 & 43.36 \\
         & \modelname & \bf 31.24 & \bf 48.61 \\
        \bottomrule
    \end{tabular}
    \caption{Abstractiveness scores as calculated by \citet{dreyer-etal-2023-evaluating} and abstractiveness-adjusted FactCC.}
    \label{tab:mint_scores}
\end{table}

\subsection{Ablation Study}
\label{sec:ablation}

In order to justify the modification made in the 
candidate ranking process of \modelname, we compute baselines that highlight the importance of each individual component. We perform the following studies using PEGASUS fine-tuned on XSUM. 

\stitle{Candidate Selecting Process} As explained in \Cref{sec:approach} we restrict the number of candidates summaries in-order to maintain a class \textit{balanced} set. We relax this constraint by simply scoring \textit{all} the candidate summaries using FactCC. This is represented by \modelname - w/o select. in \Cref{tab:ablation}. We can observe that this process leads to improved model factuality but still falls far short of the main approach by 4 points. Hence highlighting the advantage of focusing on generating quality training data.

\stitle{Dual Scoring Technique} To understand the importance of using ROUGE to select the top candidates from both factual and non-factual sets, we ablate this step by selecting the top factual and non-factual summaries using FactCC itself. 
This is marked as \modelname - w/o ROUGE in \Cref{tab:ablation}. Although the gains from this model on factuality are almost the same as \modelname, it negatively affects the ROUGE score.

\begin{table}[t!]
    \centering
    \small
    \begin{tabular}{l|c|c}
    \toprule
    \bf Model & \bf R-L & \bf FactCC \\ 
    \midrule
    PEGASUS & 39.26 & 24.33 \\
    \midrule
    \modelname - w/o select. & 38.32 & 26.38  \\ 
    \modelname - w/o ROUGE & 38.34 & 29.83 \\ 
    \midrule
    \modelname & \bf 39.45 & \bf 30.48 \\

    \bottomrule
    \end{tabular}
    \caption{Evaluation results for the various baseline models in \cref{sec:ablation}. We can observe that both the components in the ranking strategy is required in order to obtain maximum benefits from the training process.}
    \label{tab:ablation}
\end{table}

\section{Related Work}
Factual consistency in abstractive summarization has garnered much attention recently \cite{goyal-durrett-2020-evaluating,zhu-etal-2021-enhancing}.
Existing works have explored improving factual consistency during fine-tuning, inference, and pre-training stages, respectively.
For factual fine-tuning, works have applied contrastive learning \cite{cao-wang-2021-cliff, nan-etal-2021-improving}, reinforcement learning \cite{gunasekara-etal-2021-using-question} or knowledge integration \cite{zhu-etal-2021-enhancing} to teach the model identify summaries of high factual consistency while \citet{wan-bansal-2022-factpegasus} modify the pretraining process to introduce factuality-awareness. Several works have also improved summary factuality through post-processing in inference, such as correcting errors and re-ranking by factual scores \cite{cao-etal-2020-factual,dong2020multi,balachandran-etal-2022-correcting,chen2021improving,zhu-etal-2021-enhancing}.
Our work differs from the aforementioned works as we improve both factuality and summary quality, unlike other methods, which often sacrifice one for the other. 



\section{Conclusion}
We present \modelname (\textbf{E}ffective \textbf{Fact}ual \textbf{Sum}marization), a candidate summary generation and ranking technique for contrastive summarization training, which helps make models more faithful without adversely affecting summary quality. Results show that this simple, yet effective method can achieve consistent gains on both factuality and similarity-based metrics without negatively affecting the degree of abstractiveness. We hope that our findings will encourage future research on factuality-consistent summarization to focus more on the tradeoffs between summary quality and factuality.

\section*{Acknowledgement}

We appreciate the reviewers for their insightful
comments and suggestions. We would also like to thank Raj Dabre and Sumanth Doddapaneni for their feedback on the initial versions of the work.
Tanay Dixit was supported by the NSF REU Site Grant 2051101.
Fei Wang was supported by the Annenberg Fellowship at USC.
Muhao Chen was supported by the NSF Grant IIS 2105329, by Air Force Research Laboratory under
agreement number FA8750-20-2-10002, by an Amazon Research Award and a Cisco Research Award.
Computing of this work was partly supported by a subaward of NSF Cloudbank 1925001 through UCSD.

\section*{Limitations}
While our approach helps train factuality-aware summarization models, it comes at an additional computation cost. It takes 3X time to train compared to the vanilla cross-entropy model. There is also an additional overhead computational cost in generating and scoring the candidate summaries for each article in the training dataset, but we believe that the gains justify the additional computation cost. 
Improving faithfulness in summarization models is a challenging task. Although we make improvements over prior work 
by achieving improved factuality metrics, like the compared prior works, our work has not focused on numerical consistency. This could be a meaningful research direction for follow-up work.


\bibliography{anthology,custom}
\bibliographystyle{acl_natbib}

\appendix

\input{appendix}

\end{document}

%% file: main_results.tex
\definecolor{colboxcolor}{HTML}{edfbd5}

\begin{table}[t!]
    \centering
    \small
    \setlength{\tabcolsep}{5pt}
    \begin{tabular}{ l | c c c | c c} 
        \toprule
    & \multicolumn{3}{c|}{\cellcolor[HTML]{D9EAD3}\textbf{Summ. Quality}}                                   & \multicolumn{2}{c}{\cellcolor[HTML]{FCE5CD}\textbf{Factuality}}                                                                                                              \\
 
 \multirow{-2}{*}{\textbf{Model}} & \multicolumn{1}{c}{\cellcolor[HTML]{D9EAD3}\textbf{R-1}} & \textbf{\cellcolor[HTML]{D9EAD3}R-L} & \cellcolor[HTML]{D9EAD3}\textbf{BS.} & \multicolumn{1}{c}{\cellcolor[HTML]{FCE5CD}\textbf{FactCC}} & \multicolumn{1}{c}{\cellcolor[HTML]{FCE5CD}\textbf{\begin{tabular}[c|]{@{}c@{}}DAE $\downarrow$ \end{tabular}}} \\
\midrule
 \multicolumn{6}{c}{\textit{\texttt{XSUM}}} \\
 \midrule
 PEGASUS 
& 47.07                                                     
& 39.26          
& 89.19              
& 24.33                                            
& 0.426     \\
BRIO 
& \cellcolor[HTML]{edfbd5}\underline{48.69} 
& \cellcolor[HTML]{edfbd5}\underline{40.13}
& \cellcolor[HTML]{edfbd5}\underline{90.87}
& 21.47 
& 0.452  
        \\
        \midrule
 FASum                            
& 29.72                                               
& 23.29  
& 88.57           
& \cellcolor[HTML]{edfbd5}26.08                                    
& 0.616  \\
 DAE                            
& 38.63                                                          
& 30.22   
& 88.44              
& \cellcolor[HTML]{edfbd5}26.66                                     
& 0.462    \\
 CLIFF                            
& 46.33                                                          
& 38.27            
& 88.96              
& \cellcolor[HTML]{edfbd5}24.54                                     
& \cellcolor[HTML]{edfbd5} \bf \underline{0.386}                                                                       \\
 \modelname                       
& \cellcolor[HTML]{edfbd5}\bf 47.24          
& \cellcolor[HTML]{edfbd5}\bf 39.45
& \cellcolor[HTML]{edfbd5}\textbf{89.79}     
& \cellcolor[HTML]{edfbd5}\textbf{\underline{30.48}}                                              
& \cellcolor[HTML]{edfbd5} 0.417 \\
\midrule
\midrule
\multicolumn{6}{c}{\textit{\texttt{CNN/DM}}} \\
\midrule
 BART 
& 43.04
& 39.41
& 87.21
& 49.07
& 0.049 \\
BRIO 
& \cellcolor[HTML]{edfbd5}\underline{47.53}
& \cellcolor[HTML]{edfbd5}\underline{44.02}
& \cellcolor[HTML]{edfbd5}\underline{89.12}
& 30.35
& 0.093  \\
\midrule
 FASum                            
& 40.40
& 36.97                                                         
& \cellcolor[HTML]{edfbd5}88.23             
&  \cellcolor[HTML]{edfbd5}51.17                                                      
&  0.046 \\
                                      
 CLIFF                            
& \cellcolor[HTML]{edfbd5}44.14
& \cellcolor[HTML]{edfbd5}40.72
& \cellcolor[HTML]{edfbd5}\bf 88.82
& \cellcolor[HTML]{edfbd5}51.84
& 0.047
                                                                           \\

 \modelname                       
& \cellcolor[HTML]{edfbd5}\bf 44.37                                 
& \cellcolor[HTML]{edfbd5}\bf 40.92 
& \cellcolor[HTML]{edfbd5}88.36    
& \cellcolor[HTML]{edfbd5}\textbf{\underline{60.74}}                                              
& \cellcolor[HTML]{edfbd5}\textbf{\underline{0.041}}  \\

\bottomrule
    \end{tabular}
    \caption{Results of models fine-tuned on the XSUM and CNN/DM. R-1: Rouge-1 , R-L: Rouge-L , BS: BERTScore. For DAE smaller the better the score. \colorbox{colboxcolor}{Models} perform significantly better than the PEGASUS/BART model ($p<0.05$). The best result for factuality-aware training methods is \textbf{bolded}. Overall best score per metric is \underline{underlined}.}
    \label{tab:main_results}
\end{table}

%% file: examples.tex
\begin{table*}[t!]
    \scriptsize
    \centering
    \extrarowheight=\aboverulesep
    \addtolength{\extrarowheight}{\belowrulesep}
    \aboverulesep=3pt
    \belowrulesep=3pt
    \setlength{\tabcolsep}{2pt}
    \begin{tabular}{@{}c  p{0.32\textwidth} p{0.62\textwidth}}
     \toprule
 \multicolumn{1}{c}{ \bf System} & \multicolumn{1}{c}{ \bf Summary} &  \multicolumn{1}{c}{ \bf Article} \\\midrule
     \multicolumn{1}{c}{\bf Base.} &  The number of migrants and refugees arriving on the Greek island of Lesbos has halved in the past week.
     &  
     \multirow{4}{*}{\parbox[height=1.8\textwidth]{0.62\textwidth}{Lesbos used to get more than 5,000 a day. On Monday there were just four. But with Europe's borders closed, more than 50,000 migrants remain in Greece waiting for a decision about their futures. \dots But here she is in Moria, once a transit camp for migrants, now \textbf{since the EU deal with Turkey}, a detention centre, run by central government. \dots It is another sign of how Greece was simply overwhelmed by the numbers who came, while itself in the middle of an economic crisis. Most of those who arrived before March 20, the start of the EU-Turkey agreement, are free to come and go, but cannot leave the island. Those who came after that date are locked in, waiting for a decision \dots}}
 \\\cmidrule{1-2}
    \multicolumn{1}{c}{\cellcolor{gray!25}\bf Ours} 
     & \cellcolor{gray!25} The number of migrants arriving on the Greek island of Lesbos has halved since the EU struck a deal with Turkey to stem the flow.
     &  \\
 \cmidrule{0-2}
      \multicolumn{1}{c}{\bf Base} & Goldman Sachs will no longer conduct face-to-face interviews with students applying for analyst jobs.  
       &  \multirow{2}{*}{\parbox[height=1.5\textwidth]{0.62\textwidth}{The US investment bank will switch to video interviews with first-round undergraduate candidates from next month. \dots Goldman hoped the move will allow it to find students who do not attend top-tier US universities. \dots \textbf{It will still conduct second-round interviews in person}. The shift will not affect business schools or professional hires, but is part of a broader move by Goldman to use technology in the hiring process. The new method will include structured interviews, which the bank said will allow for greater comparisons between candidates \dots}}
 \\\cmidrule{1-2}
    \multicolumn{1}{c}{\cellcolor{gray!25}\bf Ours} 
     & \cellcolor{gray!25} Goldman Sachs is changing the way it hires students. \\
    \multicolumn{1}{c}{\cellcolor{gray!25}}  & \cellcolor{gray!25}  &
     \\\cmidrule{0-2}
     \multicolumn{1}{c}{\bf Base} & The pilot of a Turkish military jet has died after it crashed in the south-west of the country, state media report.  
       &  \multirow{4}{*}{\parbox[height=1.5\textwidth]{0.62\textwidth}{The plane was flying over the Amanos Mountains in the southern province of Osmaniye on Monday when it lost radio contact, Anatolia news agency said. \dots Rescuers found the pilot's body near to the wreckage of the aircraft. Osmaniye Governor Celalettin Cerrah had earlier announced that a cockpit window and some other pieces of the aircraft had been found in the Caksir area\dots People living around the village of Yarpuz, about 25km (16 miles) north of the \textbf{Syrian border, said that they had heard a loud bang like an explosion}, according to local media A Turkish fighter jet was shot down by Syria over the Mediterranean in June 2012, after Syrian forces said it had entered the country's airspace.}}
 \\\cmidrule{1-2}
    \multicolumn{1}{c}{\cellcolor{gray!25}\bf Ours} 
     & \cellcolor{gray!25}  A Turkish air force pilot has been killed after his jet crashed near the Syrian border , officials say.
 \\
  \bottomrule
\end{tabular}
\caption{Sample summaries from PEGASUS (Base) and \modelname (Ours) on XSUM articles. The information from the article that contradicts the Base summaries is in \textbf{bold}. We can see that the outputs from our fine-tuned model not only generate faithful summaries but also capture the essential information from the article well. 
}
\label{tab:example}
\end{table*}

%% file: appendix.tex
\clearpage

\section{Additional Training Details}
All experiments were carried out using 4, 24GB NVIDIA RTX A5000 GPUs. Experiments were conducted using a private infrastructure, which has a carbon efficiency of 0.432 kgCO$_2$eq/kWh. Total emissions are estimated to be 4.84 kgCO$_2$eq of which 0 percents were directly offset. Estimations were conducted using the \href{https://mlco2.github.io/impact#compute}{MachineLearning Impact calculator} presented in \cite{lacoste2019quantifying}.

\label{sec:app_imp_details}

\paragraph{XSUM}: For every news article in XSUM, we use diverse beam search \citep{Vijayakumar_Cogswell_Selvaraju_Sun_Lee_Crandall_Batra_2018} to generate 16 summaries using fine-tuned PEGASUS\footnote{google/pegasus-xsum} and 16 summaries using CLIFF (\textit{maskrel, syslowcon, swapent} and \textit{regenrel}). We use the standard ROUGE-L\footnote{https://github.com/summanlp/evaluation/tree/master/ROUGE-RELEASE-1.5.5} implementation and for FactCC, we use the \texttt{FactCC} checkpoint from the official implementation provided by the authors\footnote{https://github.com/salesforce/factCC}. Articles for which we are unable to generate the required number of factual and non-factual summaries are discarded. In the end, our training dataset contains 145,040 data points. Choosing a bigger candidate size (>6) led to a decrease in the training dataset size as mentioned in \S\ref{sec:ranking_strat}. 
\begin{table}[h]
    \centering
    \begin{tabular}{l|l}
    \toprule
    \bf Hyperparameters & \bf Value \\
    \midrule
    \midrule
    model &  google/pegasus-xsum \\
    no. of params & 568M \\
    max learning rate & 1e-4\\
    warmup steps & 500 \\
    number of epochs & 5 \\
    per device batch size & 1 \\ 
    accumulation step & 16 \\
    margin & 0.001 \\
    max seq length  & 512 \\
    mle weight & 1 \\
    ranking weight & 10 \\
    \bottomrule
    \end{tabular}
    \caption{Hyperparameters for PEGASUS on XSUM.}
    \label{tab:xsum_hyper}
\end{table}

\paragraph{CNN/DM} For CNN/DM we follow the same process as described for XSUM, except here we use BART Large\footnote{facebook/bart-large-cnn}. For CLIFF on CNN/DM we use \textit{syslowcon\_maskrel, syslowcon, syslowcon\_swapent} and \textit{syslowcon\_regenrel} models. In the end our training dataset has 246,796 articles.

\paragraph{Training details} For training we use the Adam optimizer with linear learning rate scheduling for the model training. \Cref{tab:xsum_hyper} and \Cref{tab:cnndm_hyper} contain the best set of hyper-parameters for training PEGASUS on XSUM and BART on CNN/DM. These hyper-parameters were obtained after an extensive grid search. We perform validation after every 1600 steps and save the best model using the validation cross-entropy loss.

\begin{table}[h]
    \centering
    \begin{tabular}{l|l}
    \toprule
    \bf Hyperparameters & \bf Value \\
    \midrule
    \midrule
    model &  facebook/bart-large-cnn \\
    no. of params & 400M \\
    max learning rate & 3e-5\\
    warmup steps & 500 \\
    number of epochs & 5 \\
    per device batch size & 1 \\ 
    accumulation step & 16 \\
    margin & 0.001 \\
    max seq length  & 1024 \\
    mle weight & 0.1 \\
    ranking weight & 10 \\
    \bottomrule
    \end{tabular}
    \caption{Hyperparameters for BART on CNN/DM.}
    \label{tab:cnndm_hyper}
\end{table}

\paragraph{Decoding parameters} We follow \citet{cao-wang-2021-cliff} and use the beam search algorithm to decode summaries. For BART, we set the beam sizes as 4 on CNN/DM and a beam size of 8 is used for PEGASUS on XSUM. The additional decoding parameters are in \Cref{tab:decod_hyper}.

\begin{table}[h]
    \centering
    \begin{tabular}{ll}
    \toprule
    \bf Hyperparameters & \bf Value \\
    \midrule
    \multicolumn{2}{c}{\bf BART} \\
    \midrule
    beam size & 4 \\
    length penalty & 2 \\
    max-length & 140 \\
    min-length & 55 \\
    \midrule
    \multicolumn{2}{c}{\bf PEGASUS} \\
    \midrule
    beam size & 8 \\
    length penalty & 0.6 \\
    max-length & 62 \\
    min-length & 11 \\
    \bottomrule
    \end{tabular}
    \caption{Decoding parameters for BART and PEGASUS}
    \label{tab:decod_hyper}
\end{table}

\section{Extractiveness Results}
\label{sec:app_extr_results}
The extractivenes scores as calculated using the \textit{coverage} score defined by \citet{grusky-etal-2018-newsroom} are present in \Cref{tab:extractive_analysis_cnndm} and \Cref{tab:extractive_analysis_xsum}. Lower the score the higher the abstraction. We can observe that \modelname achieves a lower abstraction level than CLIFF on both the datasets.
\begin{table}[h]
\centering
\begin{tabular}{l | c}
\toprule
\textbf{Model} & \textbf{Abstractiveness ($\downarrow$)}  \\
\midrule
\midrule
Reference                          & \textbf{0.666}                      \\
Pegasus                            & 0.735                  \\
CLIFF                              & 0.759                  \\
\modelname                         & 0.720                  \\
\bottomrule
\end{tabular} \\
    \caption{Extractivness analysis for XSUM}
    \label{tab:extractive_analysis_xsum}
\end{table}

 \begin{table}[h]
\centering
\begin{tabular}{l | c}
\toprule
\textbf{Model} & \textbf{Abstractiveness ($\downarrow$)}  \\
\midrule
\midrule
Reference                          & \textbf{0.880}                      \\
BART                            & 0.991                  \\
CLIFF                              & 0.989                  \\
\modelname                         & 0.979                  \\
\bottomrule
\end{tabular} \\
    \caption{Extractivness analysis for CNN/DM}
    \label{tab:extractive_analysis_cnndm}
\end{table}

\section{Generated outputs}

More examples generated outputs by \modelname on different backbones and raw documents are in \Cref{tab:app_xsum,tab:app_cnndm}.

\label{sec:app_outputs}
\begin{table*}[t!]
    \scriptsize
    \centering
    \extrarowheight=\aboverulesep
    \addtolength{\extrarowheight}{\belowrulesep}
    \aboverulesep=2pt
    \belowrulesep=2pt
    \setlength{\tabcolsep}{2pt}
    \begin{tabular}{@{}c  p{0.25\textwidth} p{0.6\textwidth}}
     \toprule
 \multicolumn{1}{c}{ \bf System} & \multicolumn{1}{c}{ \bf Summary} &  \multicolumn{1}{c}{ \bf Article} \\\midrule
     \multicolumn{1}{c}{\bf Base.} &  The number of migrants and refugees arriving on the Greek island of Lesbos has halved in the past week.
     &  
     \multirow{4}{*}{\parbox[height=1.8\textwidth]{0.6\textwidth}{Lesbos used to get more than 5,000 a day. On Monday there were just four. But with Europe's borders closed, more than 50,000 migrants remain in Greece waiting for a decision about their futures. \dots But here she is in Moria, once a transit camp for migrants, now \textbf{since the EU deal with Turkey}, a detention centre, run by central government. \dots It is another sign of how Greece was simply overwhelmed by the numbers who came, while itself in the middle of an economic crisis. Most of those who arrived before March 20, the start of the EU-Turkey agreement, are free to come and go, but cannot leave the island. Those who came after that date are locked in, waiting for a decision \dots}}
 \\\cmidrule{1-2}
    \multicolumn{1}{c}{\cellcolor{gray!25}\bf Ours} 
     & \cellcolor{gray!25} The number of migrants arriving on the Greek island of Lesbos has halved since the EU struck a deal with Turkey to stem the flow .
     &  \\
 \cmidrule{0-2}
      \multicolumn{1}{c}{\bf Base} & Hundreds of eggs from two rare bird species have been stolen. 
       &  \multirow{2}{*}{\parbox[height=1.5\textwidth]{0.6\textwidth}{The Mediterranean gull and black-headed gull eggs were illegally harvested from from islands in Poole Harbour, Dorset.\dots Natural England is urging any restaurants or pubs to ask to see a valid licence before buying eggs to prepare in meals. Birds of Poole Harbour had been surveying a group of islands in the harbour when the theft was discovered. Mediterranean gulls are classified as a Schedule One species, meaning anyone disturbing their nests must have a special licence. Paul Morton, who runs the charity, said Mediterranean gulls' eggs were not approved for human consumption, and could be a "health issue". "I'm distraught, really. To see the taking of hundreds and hundreds of eggs from an important colony is quite sickening," he said. Mr Moreton said there had been previous convictions for egg poaching in the last 10 or 15 years...}}
 \\\cmidrule{1-2}
    \multicolumn{1}{c}{\cellcolor{gray!25}\bf Ours} 
     & \cellcolor{gray!25} Hundreds of gull eggs have been stolen from a protected colony.\\
    \multicolumn{1}{c}{\cellcolor{gray!25}}  & \cellcolor{gray!25}  & \\
    \multicolumn{1}{c}{\cellcolor{gray!25}}  & \cellcolor{gray!25}  &
     \\\cmidrule{0-2}
     \multicolumn{1}{c}{\bf Base} & A volcano in western Indonesia has erupted for the second time in two years, killing at least 11 people, officials say.  
       &  \multirow{4}{*}{\parbox[height=1.5\textwidth]{0.6\textwidth}{The victims were farming in an area that was declared unsafe because of its close proximity to Mount Sinabung. The volcano was still spewing ash on Sunday, hampering rescue operations. More than a dozen people were killed when it erupted in 2014. It also erupted in 2010, after having been dormant for 400 years. Rescue teams are still scouring the area, looking for more victims who may have been killed or badly burned by the hot gas and ash clouds released in the eruption. Rescue teams were searching homes and farms in the village of Gamber, which was also evacuated in 2014. What causes volcanoes? The 2,460-metre (8,070 foot) tall volcano is among the country's most active. Indonesia, located on the Pacific Ring of Fire, has more than 120 active volcanoes.}}
 \\\cmidrule{1-2}
    \multicolumn{1}{c}{\cellcolor{gray!25}\bf Ours} 
     & \cellcolor{gray!25}  At least 11 people have been killed after a volcano on the Indonesian island of Sumatra erupted , officials say .
     \\\cmidrule{0-2}
     \multicolumn{1}{c}{\bf Base} & The SNP and Labour have won seats on Edinburgh Council in two by-elections. 
       &  \multirow{4}{*}{\parbox[height=1.5\textwidth]{0.6\textwidth}{It was the first time the Single Transferable Vote (STV) system had been used to select two members in the same ward in a by-election. The SNP topped the vote in the Leith Walk by-election, while Scottish Labour won the second seat from the Greens. The by-election was called after Deidre Brock of the SNP and Maggie Chapman of the Scottish Greens stood down.\dots The turnout for the by-election was 25.1\%. The SNP also held the Midlothian West seat on Midlothian Council with a swing of 6.3\% from Labour. The party's Kelly Parry secured 1,540 votes, ahead of Labour's Ian Miller on 945 votes. The by-election was called after Owen Thompson was elected as SNP MP for the Midlothian constituency.}}
 \\\cmidrule{1-2}
    \multicolumn{1}{c}{\cellcolor{gray!25}\bf Ours} 
     & \cellcolor{gray!25} A by-election has been held in Edinburgh to fill two seats on the city council . & \\
    \multicolumn{1}{c}{\cellcolor{gray!25}}  & \cellcolor{gray!25}  &
 \\
  \bottomrule
\end{tabular}
\vspace{-0.5em}
\caption{Sample summaries from PEGASUS (Base) and \modelname (Ours) on XSUM articles.}
\vspace{-1.5em}
\label{tab:app_xsum}
\end{table*}

\begin{table*}[t!]
    \scriptsize
    \centering
    \extrarowheight=\aboverulesep
    \addtolength{\extrarowheight}{\belowrulesep}
    \aboverulesep=1pt
    \belowrulesep=1pt
    \setlength{\tabcolsep}{2pt}
    \begin{tabular}{@{}c  p{0.32\textwidth} p{0.62\textwidth}}
     \toprule
 \multicolumn{1}{c}{ \bf System} & \multicolumn{1}{c}{ \bf Summary} &  \multicolumn{1}{c}{ \bf Article} \\\midrule
 \cmidrule{0-2}
      \multicolumn{1}{c}{\bf Base} & Video shows the lions interacting with the visitors who stand inside a metal cage attached to a car. The video was captured by ekant veer, 35, an associate professor at the university of canterbury. A number of the lions are fed directly through the metal bars, while others receive meat dropped from the back of the cage.  
       &  \multirow{2}{*}{\parbox[height=1.5\textwidth]{0.62\textwidth}{visitors to a wildlife park in new zealand got to encounter a pride of lions up-close and personal. filmed at the orana wildlife park  the countrys only open-range zoo  the video shows the lions interacting with the visitors who stand inside a metal cage attached to a car. the video, which was captured by ekant veer, 35, an associate professor at the university of canterbury, also shows the lions scaling the cage and eating meat through its bars.\dots as the keeper speaks, the lion licks at a piece of meat that is held up against the bars as another lion walks across the roof of the cage. looking down at the people below, the lion wanders around as if deciding who it would like to make its prey before staring down the lens of the camera. set tongues wagging! one of the lions notices meat and begins sticking out its tongue in the hope of being fed. a lion stands next to one of the keepers and its large paw is the same size as the lady's head. the people inside can be seen recording the many lions from their phones, while another  with paws the same size as the keepers head  holds itself up against the cage and chews on some meat. later in the video people can be seen pointing out the various felines as a keeper moves her hand along the cage, instigating the lion to follow. still frames capture a lion standing up against the side of the cage alongside the keeper  its power and size is plain to see.\dots orana wildlife trust. located on the outskirts of christchurch, the wildlife park is unique in that the people are caged in order to view the animals, not the other way around.}}
 \\\cmidrule{1-2}
    \multicolumn{1}{c}{\cellcolor{gray!25}\bf Ours} 
     & \cellcolor{gray!25} the video was filmed at the orana wildlife park in new zealand , the country 's only open-range zoo . the video shows the lions interacting with the visitors who stand inside a metal cage attached to a car . a number of the lions are fed directly through the metal bars , while others receive meat dropped from the back of the cage . \\
    \multicolumn{1}{c}{\cellcolor{gray!25}}  & \cellcolor{gray!25}  &
     \\\cmidrule{0-2}
     \multicolumn{1}{c}{\bf Base} & Taxpayers are having to find 11billion a year to top up the wages of millions of people working in supermarkets and other low paid jobs. Money is paid to some 5.2million workers in the form of tax credits and other benefits. Total amount of benefits paid to staff at some companies exceeds what the firms pay in corporation tax. 
       &  \multirow{4}{*}{\parbox[height=1.5\textwidth]{0.62\textwidth}{taxpayers are having to find 11billion a year to top up the wages of millions of people working in supermarkets and other low paid jobs. the money, which amounts to a massive public subsidy for the companies involved, is paid to some 5.2million workers in the form of tax credits and other benefits. \dots the charity is campaigning for the adoption of the living wage - 9.15 an hour in london and 7.85 for the rest of the uk - across both the public and private sector. it estimates this would reduce the need for in-work benefits by 6.7bn a year, which would make a massive dent in the 12billion reduction in welfare spending which the conservatives say is necessary. the current minimum wage for those over 21 is 6.50 an hour and will rise to 6.70 in october, da and sainsburys posted combined profits of 3.9bn last year, but between them cost the taxpayer more than 750m in benefits paid to their staff. tesco paid 519m in tax but received 364m in public subsidy for its 209,000 low-paid workers. asda spent 150m in tax but its 120,000 low-paid workers received 221m in benefits. \dots thesupermarkets said they paid above the minimum wage of 6.50 an hour for those aged over 21, regularly reviewed pay and gave employees benefits such as staff discounts. asda, which is part of the us retail goliath walmart, said pay and benefits should be considered in the round. in the usa, it is estimated that walmarts low-wage workers cost u.s. taxpayers an estimated \$6.2 billion (4.2bn) in public assistance including food stamps, medicaid and subsidised housing. \dots}}
 \\\cmidrule{1-2}
    \multicolumn{1}{c}{\cellcolor{gray!25}\bf Ours} 
     & \cellcolor{gray!25} Taxpayers are having to find 11billion a year to top up the wages of millions of people working in supermarkets and other low paid jobs. Money is paid to some 5.2million workers in the form of tax credits and other benefits. Total amount of benefits paid to staff at some companies exceeds what the firms pay in corporation tax.
 \\
  \bottomrule
\end{tabular}
\caption{Sample summaries from BART Large (Base) and \modelname (Ours) on CNN/DM articles.}
\label{tab:app_cnndm}
\end{table*}